\pdfoutput=1

\documentclass[journal,twoside]{IEEEtran}

\IEEEoverridecommandlockouts                              

\usepackage{times}

\usepackage[nolist]{acronym}
\usepackage{amsmath}
\usepackage{placeins}
\usepackage[export]{adjustbox}
\usepackage{amssymb}
\usepackage{bm}
\usepackage{booktabs}
\usepackage{balance}
\usepackage{csquotes}
\usepackage{color}
\usepackage{colortbl}
\usepackage{dsfont}
\usepackage{esvect}
\usepackage{float}
\usepackage{graphicx}
\usepackage{import}
\usepackage{indentfirst}
\usepackage{mathtools}
\usepackage{multicol}
\usepackage{multirow}
\usepackage{flushend}
\usepackage{outline}
\usepackage{paralist}
\usepackage{pifont}
\usepackage{siunitx}
\usepackage{stmaryrd}
\usepackage{systeme}
\usepackage{soul}
\usepackage{optidef}
\usepackage{url}
\usepackage{tikz}
\usepackage{tabularx}
\usepackage{verbatim}
\usepackage{cite}
\usepackage{hyperref}
\usepackage{eso-pic}
\AddToShipoutPictureBG*{\AtPageLowerLeft{\raisebox{0.22in}{\makebox[\paperwidth]{\parbox[b]{\textwidth}{\centering\scriptsize\copyright~2026 IEEE. Personal use of this material is permitted. Permission from IEEE must be obtained for all other uses, in any current or future media, including reprinting/republishing this material for advertising or promotional purposes, creating new collective works, for resale or redistribution to servers or lists, or reuse of any copyrighted component of this work in other works.}}}}}
\usetikzlibrary{tikzmark}
\usepackage{xcolor,pifont}

\newcommand{\cmark}{\color{green}\ding{51}}%
\newcommand{\xmark}{\color{red}\ding{55}}%

\hypersetup{
  hidelinks,
  pdfauthor={Martin Peticco, Gabriella Ulloa, John Marangola, Nitish Dashora, Pulkit Agrawal},
  pdftitle={DexWrist: A Robotic Wrist for Constrained and Dynamic Manipulation},
  pdfsubject={Robotic Wrist},
  pdfkeywords={Dexterous Manipulation; Mechanism Design of Manipulators; Telerobotics and Teleoperation}
}

\makeatletter
\def\bstctlcite{\@ifnextchar[{\@bstctlcite}{\@bstctlcite[@auxout]}}
\def\@bstctlcite[#1]#2{\@bsphack
  \@for\@citeb:=#2\do{%
    \edef\@citeb{\expandafter\@firstofone\@citeb}%
    \if@filesw\immediate\write\csname #1\endcsname{\string\citation{\@citeb}}\fi}%
  \@esphack}
\makeatother

\begin{document}

\title{DexWrist: A Robotic Wrist for Constrained and Dynamic Manipulation}

\author{Martin Peticco, Gabriella Ulloa, John Marangola, Nitish Dashora, and Pulkit Agrawal%
\thanks{Manuscript received: April 16, 2026; Revised: July 18, 2026; Accepted: September 1, 2026.}%
\thanks{This paper was recommended for publication by Editor Clement Gosselin upon evaluation of the Associate Editor and Reviewers' comments.}%
\thanks{This work was supported by ONR MURI (grant N00014-22-1-2740), Sony, and Toyota Research Institute.}%
\thanks{The authors are with the Improbable AI Lab, Massachusetts Institute of Technology, Cambridge, MA 02139 USA (e-mail: \{mpeticco, gulloa, jmgola, dashora, pulkitag\}@mit.edu).}%
\thanks{Videos and details: \href{https://martinpeticco.com/dexwrist}{martinpeticco.com/dexwrist}}%
\vspace{-1mm}}

\markboth{IEEE ROBOTICS AND AUTOMATION LETTERS. PREPRINT VERSION. ACCEPTED SEPTEMBER, 2026}{Peticco \MakeLowercase{\textit{et al.}}: DexWrist: A Robotic Wrist for Constrained and Dynamic Manipulation}
\maketitle
\bstctlcite{DexWrist:BSTcontrol}
\begin{figure*}[ht]
    \centering
    \includegraphics[width=1.0\linewidth]{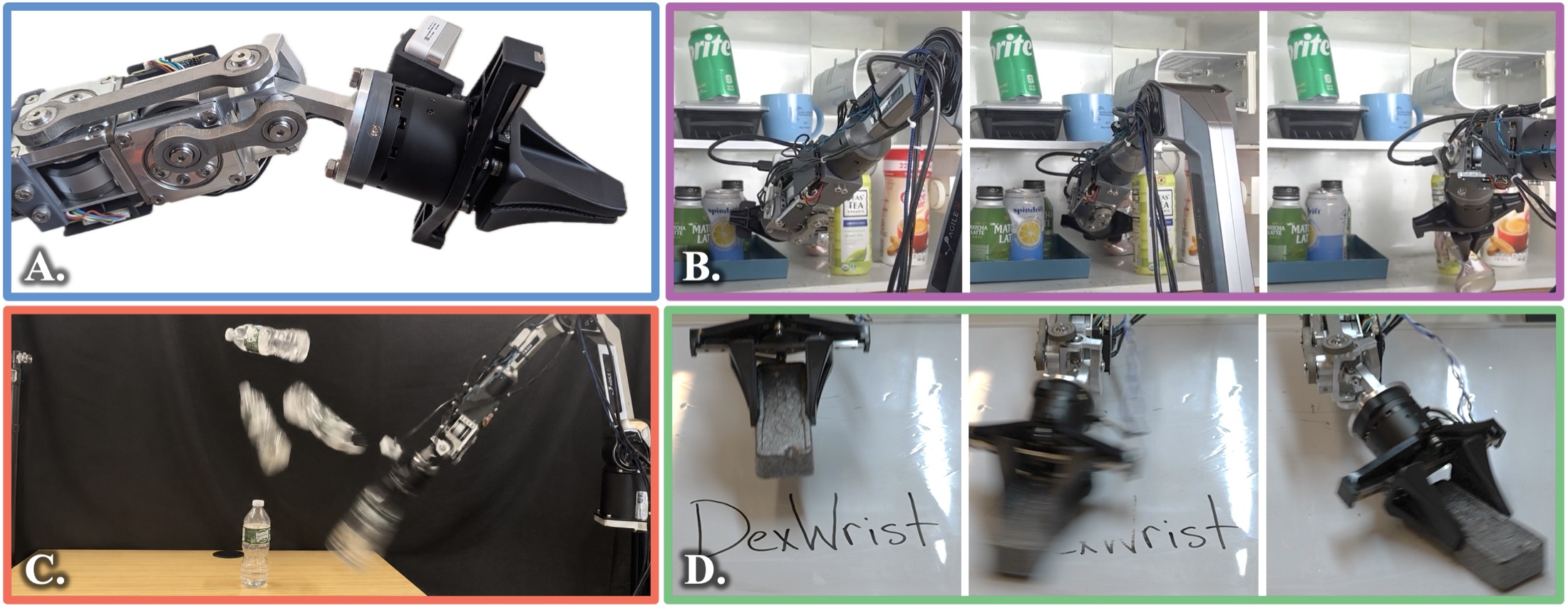}
    \caption{\textit{DexWrist}, a robotic wrist for constrained and dynamic manipulation. \textbf{A:} \textit{DexWrist} with an AgileX gripper. \textbf{B:} Learned constrained-space task: picking from a cluttered fridge. \textbf{C:} Dynamic bottle flip (pre-planned). \textbf{D:} Learned contact-rich task: wiping a whiteboard.}
    \label{fig:wristhero}
    \vspace{-2mm}
\end{figure*}

\begin{abstract}
Development of dexterous manipulation hardware has primarily focused on hands and grippers. However, these end-effectors are often paired with bulky and highly stiff wrists that limit performance in human environments. More recent designs have adopted backdrivable actuation, but are still difficult to model and control due to coupled kinematics or high mechanical inertia from heavy links. We present \textit{DexWrist}, a compact robotic wrist combining quasi-direct-drive actuation with a decoupled parallel kinematic mechanism to advance manipulation in highly constrained environments and enable dynamic, contact-rich tasks. It delivers \bm{$3.75\pm0.05$}~Nm rated torque, \bm{$0.33\pm0.06$}~Nm backdrive torque, \bm{$10.15\pm1.34$}~Hz torque bandwidth, \bm{$\pm40^\circ$} ROM in both DOFs, and a diagonal velocity-constraint Jacobian (one-to-one motor-to-DOF mapping) in a 0.97\,kg package. In practice, these properties increase workspace in clutter and stabilize contact without finely tuned admittance control. We evaluate \textit{DexWrist} as a drop-in upgrade in simulation and on three robot arms across constrained and contact-rich tasks. In learned policy evaluations on the AgileX PiPER and UR3e, \textit{DexWrist} achieved \textbf{50--76\%} relative improvements in success rate and reduced autonomous task completion times by \textbf{3--5}$\times$; on a torque-controlled Franka FR3, where a strong joint-impedance baseline already succeeds, it still completed the task 1.4$\times$ faster.
\end{abstract}

\vspace{-2mm}

\IEEEpeerreviewmaketitle


\vspace{-1mm}
\section{Introduction}

\IEEEPARstart{W}{hile} robotic manipulation has advanced rapidly using machine learning and commodity hardware, current systems perform quasi-static tasks in clutter-free environments with open workspaces (e.g., empty tables and fridges), as evidenced in large-scale manipulation datasets~\cite{open_x_embodiment_rt_x_2023}. The real world is highly cluttered, imposing workspace constraints on robots that lack the compliance and force sensing to make safe contact, risking damage to the robot and the environment. Many tasks, such as wiping and cooking, also require high speed. Furthermore, most learning-based systems rely on expensive real-world data collection, making fast demonstration speed and high reliability critical. We posit that many of these shortcomings stem from the wrists on current robots, and can be mitigated by a drop-in hardware upgrade.  

A majority of current robotic systems, such as arms by Franka~\cite{noauthor_franka_nodate} and Universal Robots~\cite{universal_robots_products}, rely on parallel-jaw grippers mounted on 6- or 7-DOF arms whose wrists present several issues: (i) rigid joints with high gear reductions are slow and hard to backdrive, resulting in poor adaptation to external forces and precluding dynamic tasks~\cite{cheetah_actuators}; (ii) the joints are often too large for cluttered spaces~\cite{negrello_compact_2019}; (iii) the joints are connected end-to-end serially, making small end-effector (EE) reorientations require large arm motions, further preventing operation in cluttered spaces~\cite{negrello_compact_2019}; and (iv) parallel designs have coupled kinematics, producing a non-diagonal constraint Jacobian that increases modeling and control complexity~\cite{bajajparallelwrist}. Together, these factors limit the speed, workspace, and contact handling of current systems.

Humans, in contrast, can reorient their hands extensively in tight spaces because their wrist DOFs are co-located near the hand, reducing the motion needed to accomplish tasks. Human joints are also fast and tolerant of repeated contact~\cite{petric_hammering_2017}. In fact, studies have found that wrist dexterity with a simple end-effector outperforms a highly dexterous end-effector paired with a limited wrist~\cite{bajaj_state_2019}.

We present \textit{DexWrist} (Fig. \ref{fig:wristhero}), a compact 2-DOF robotic wrist that combines a decoupled parallel kinematic mechanism (PKM) with quasi-direct-drive (QDD) actuation. The PKM co-locates two rotational DOFs about a single point, keeping the wrist small and reducing arm motions. Its diagonal velocity-constraint Jacobian gives a one-to-one motor-to-DOF mapping that reduces modeling and control complexity. QDD actuation permits low mechanical impedance, enabling fast motions and stable contact without complex admittance control. \textit{DexWrist} is a drop-in pitch--yaw upgrade. Throughout this paper, pitch denotes flexion/extension (F/E), yaw denotes radial/ulnar (R/U), and roll denotes pronation/supination (P/S), the last of which is typically available upstream on all 6- and 7-DOF arms. Our contributions are:

\begin{compactitem}
\item A compact, robust, decoupled PKM wrist design with a co-located pitch--yaw center and one-to-one motor-to-DOF mapping.
\item A requirements-driven validation of torque, backdrive torque, bandwidth, speed, ROM, size, and weight.
\item A system-level evaluation demonstrating these properties carry downstream: 88\% larger workspace in simulation, \textbf{1.3--2.2$\times$} faster teleoperated demonstrations on the AgileX and UR3e, and improved autonomous policy performance: \textbf{50--76\%} relative increases in success rate and \textbf{3--5$\times$} faster autonomous completion on the AgileX and UR3e, and a 1.4$\times$ speedup on a torque-controlled Franka.
\end{compactitem}

The remainder of the paper covers related work (Sec.~II), design and validation (Sec.~III), system-level setup and results (Secs.~IV--V), and discussion (Sec.~VI).


\section{Prior Work}
\label{sec:priorworks}

Robotic wrists group broadly by kinematic architecture and actuation strategy; we review each category and identify the gaps that motivate our design.

Most commercially available robot arms, including the UR series~\cite{universal_robots_products}, Franka~\cite{noauthor_franka_nodate}, and AgileX PiPER~\cite{noauthor_piper_nodate}, use a serial wrist with single-DOF joints connected end-to-end along the kinematic chain. The harmonic drives common in these platforms (typically 80:1--160:1 reduction) make the wrist slow and poorly suited to contact-rich tasks due to their low backdrivability~\cite{cheetah_actuators}. Even with highly backdrivable actuators, if the joints are spaced apart (like the AgileX) rather than co-located as in the human wrist~\cite{bajaj_state_2019}, small end-effector reorientations will require large arm motions in constrained spaces~\cite{negrello_compact_2019}.

A simple approach to co-locating DOFs is a universal-joint (U-joint) arrangement. Platforms such as the Unitree H1-2~\cite{unitree_h1_2}, GALAXEA R1~\cite{galaxea_r1_manual}, and OpenArm~\cite{openarm} mount the first motor stationary in the forearm and rotate the second motor in its entirety through a linkage. This yields a one-to-one motor-to-DOF mapping, but moving motor mass raises distal inertia, reducing mechanical bandwidth and limiting dynamic tasks.

Some works address contact stability through material compliance or stiffness modulation. Milazzo et al.~\cite{milazzovariablewrist} use four motors and nonlinear elastic elements to vary stiffness up to 3$\times$, at the cost of weight and complexity. Sun et al.~\cite{sunmodularwrist} achieve 232$\times$ stiffness variation in a 55~mm, 200~g module, but its sheet-type flexure limits transmissible torque. Baggetta et al.~\cite{BaggetaCableWrist} use cable-driven flexural pivots with wide ROM, but cables introduce friction and hysteresis. BiFlex~\cite{biflex} uses a passive buckling honeycomb, but provides no active reorientation and only a 500~g payload. These designs sacrifice torque capacity, active control authority, or both.

Parallel kinematic mechanisms (PKMs) permit tighter motions by co-locating the output DOFs through linkages. In coupled designs like the Omni-Wrist~\cite{sofka_omni-wrist_2006}, Carpal wrist~\cite{alvarez_design_2019}, Damerla prosthetic wrist~\cite{damerla_design_2022}, and ByteWrist~\cite{bytewrist}, the constraint Jacobian is non-diagonal: actuators do not map one-to-one with output DOFs, requiring multi-input coordination and preventing direct per-axis torque attribution~\cite{bajajparallelwrist}. A decoupled parallel mechanism~\cite{caron1997agileeye2dof} provides one-to-one motor-to-DOF mapping with co-located rotation. The Negrello wrist~\cite{negrello_compact_2019} employs one, but relies on heavily geared (350:1) actuation, making it heavier and lower-bandwidth while limiting its use for dynamic, contact-rich tasks.

As Table~\ref{tab:wrist_comparison} summarizes, no existing wrist simultaneously provides a compact co-located rotation center, a one-to-one motor-to-DOF mapping, low-impedance (quasi-direct-drive) actuation, and low distal inertia. \textit{DexWrist} achieves this by pairing a 2-(R,~RR) decoupled PKM with custom quasi-direct-drive actuators housed entirely in the forearm.

\begin{table}[t]
  \centering
  \footnotesize
  \setlength{\tabcolsep}{4.5pt}
  \renewcommand{\arraystretch}{1.0}
  \caption{Representative wrist architectures across the properties needed for compact, contact-rich, dynamic manipulation ({\cmark}~favorable, {\xmark}~unfavorable). A non-elastic transmission has no series-elastic element or flexure, leaving endpoint stiffness to the controller. The last two columns are the inertial contributions to endpoint impedance: reflected rotor inertia and distal moving mass.}
  \label{tab:wrist_comparison}
  \scriptsize
  \setlength{\tabcolsep}{1.5pt}
  \begin{tabular}{@{}l ccccc@{}}
    \toprule
      & & & & \multicolumn{2}{c@{}}{Endpoint impedance:} \\
    \cmidrule(l){5-6}
    Architecture
      & \shortstack[c]{Co-\\located}
      & \shortstack[c]{De-\\coupled}
      & \shortstack[c]{Non-Elastic\\transmission}
      & \shortstack[c]{Reflected\\inertia}
      & \shortstack[c]{Distal\\inertia} \\
    \midrule
    Serial (UR3e, Franka)~\cite{universal_robots_products,noauthor_franka_nodate}
      & \xmark & \cmark & \cmark & high~({\xmark}) & high~({\xmark}) \\
    U-joint (OpenArm)~\cite{openarm,damiao_j4310}
      & \cmark & \cmark & \cmark & low~({\cmark}) & high~({\xmark}) \\
    Coupled PKM (ByteWrist)~\cite{bytewrist}
      & \cmark & \xmark & \cmark & low~({\cmark}) & high~({\xmark}) \\
    Compliant / VS (Milazzo)~\cite{milazzovariablewrist}
      & \xmark & \xmark & \xmark & high~({\xmark}) & low~({\cmark}) \\
    Decoupled PKM (Negrello)~\cite{negrello_compact_2019}
      & \cmark & \cmark & \xmark & high~({\xmark}) & low~({\cmark}) \\
    \textbf{\textit{DexWrist} (ours)}
      & \cmark & \cmark & \cmark & low~({\cmark}) & low~({\cmark}) \\
    \bottomrule
  \end{tabular}
  \vspace{-2mm}
\end{table}

\section{DexWrist Design and Validation}
\label{sec:hardware}
This section defines \textit{DexWrist}'s mechanical and actuation requirements, the design decisions that meet them, and the experiments validating them. Table \ref{tab:functional_reqs} summarizes the results.

\begin{table}[ht]
  \setlength{\tabcolsep}{4pt}
  \renewcommand{\arraystretch}{0.92}
  \caption{Functional requirements and measured performance. Spec denotes the design target derived from the cited work. ROM spec denotes minimum coverage and can be exceeded.}
  \label{tab:functional_reqs}

  \begin{tabular*}{\columnwidth}{@{\extracolsep{\fill}}lccc}
    \toprule
    Requirement$^{\ddagger}$ & Spec & Ours & Meets \\
    \midrule
    Rated torque (Nm)                     & $\ge 3$~\cite{pando_characterization_2013}  & $3.75\pm0.05$     & \cmark \\
    Backdrive torque (Nm)                 & $\le 0.4$~\cite{pando_characterization_2013} & $0.33\pm0.06$     & \cmark \\
    Hardstop load cap. (Nm)$^{\dagger}$                & $\ge 9$~\cite{nasa_std_5017b}     & $\ge14$                 & \cmark \\
    Axial load cap. (kg)$^{\dagger}$                & $\ge 15$~\cite{nasa_std_5017b}    & $\ge100$                 & \cmark \\
    No-load speed (rpm)                 & $\ge 53.3$~\cite{vaisman_comparative_2013}     & $96.6\pm9.4$      & \cmark \\
    Torque BW (Hz) @ 3.75\,Nm                  & $\ge 10$~\cite{forgaard_voluntary_2015}     & $10.15\pm1.34$    & \cmark \\
    Angular precision (deg)                & $\le 3.47$~\cite{holman_accuracy_2020}     & 1.65              & \cmark \\
    F/E ROM (deg)                       & $[-40^\circ,40^\circ]$~\cite{ryu_functional_1991}        & $[-40^\circ,40^\circ]$        & \cmark \\
    R/U ROM (deg)                       & $[-10^\circ,30^\circ]$~\cite{ryu_functional_1991}        & $[-40^\circ,40^\circ]$        & \cmark \\
    Width (mm)                          & $\le 61.4$~\cite{anthropometry}              & 64                & \xmark \\
    Height (mm)                         & $\le 61.4$~\cite{anthropometry}              & 66.5              & \xmark \\
    Length (mm)                         & $\le 195.5$~\cite{anthropometry}          & 178.2             & \cmark \\
    Weight (kg)                         & $\le 1.0$~\cite{robotiq_2f}              & 0.97              & \cmark \\
    \bottomrule
  \end{tabular*}
  \footnotesize{$^{\dagger}$ Load capacities are \emph{tested-to} values, not failure limits. $^{\ddagger}$ \textit{DexWrist} provides F/E and R/U; P/S is assumed upstream.}
  \vspace{-2.5mm}
\end{table}

\subsection{Functional Requirements}
\label{sec:frs}
The human wrist serves as our design reference. It provides two orientation DOFs via a condyloid/ellipsoidal joint with approximately co-located axes near the hand, while a third DOF (pronation/supination) originates upstream in the forearm~\cite{bajaj_state_2019}. We use human performance data from the literature to derive the targets below.

\subsubsection{Torque, Load Capacity, and Backdrivability}
\begin{figure}[t!]
    \centering
    \includegraphics[width=1.0\linewidth]{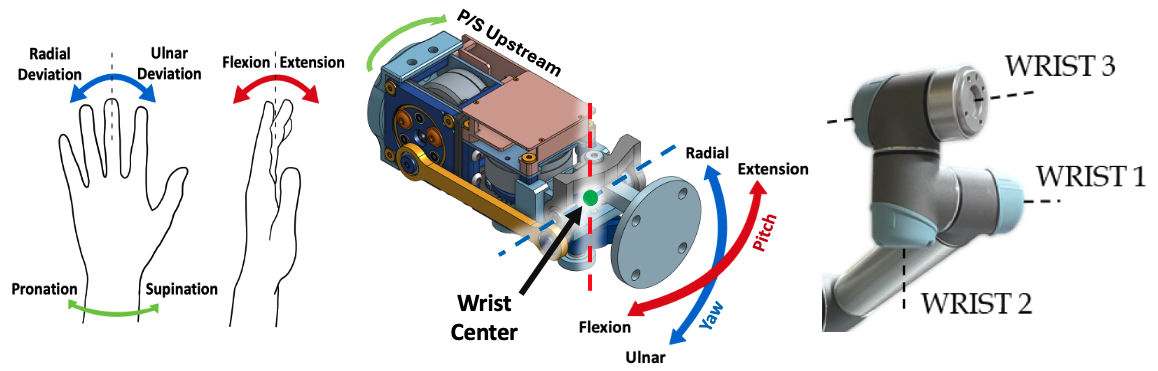}
    \caption{\textbf{Left:} Human wrists have 3 DOFs: F/E, R/U, and P/S. \textbf{Middle:} \textit{DexWrist} provides pitch (F/E) and yaw (R/U); with the arm's upstream roll (P/S), all three axes intersect at a point. \textbf{Right:} The UR3e's wrist axes do not intersect at a point.}
    \label{fig:wristDOFs}
    \vspace{-2.5mm}
\end{figure}

93\% of activities of daily living (ADL) can be completed with a torque of 3 Nm in both the R/U and F/E directions (Fig.~\ref{fig:wristDOFs})~\cite{pando_characterization_2013}. We do not target P/S torque, as most 6- and 7-DOF robot arms provide this DOF.

For static loads at the output flange, we follow NASA hardstop guidance~\cite{nasa_std_5017b} with a 3.0 factor of safety over rated torque, giving 9~Nm at the hardstops, and target a 15~kg axial load for UR5e-class payloads.

Low-impedance actuation lets the wrist conform to the environment while sustaining unexpected impacts. We interpret this as a 5~N maximum backdrive force~\cite{pando_characterization_2013} at 70~mm from the wrist center, i.e., $\le 0.35$~Nm, rounded to 0.4~Nm.

\subsubsection{Speed, Bandwidth, Range, and Precision}
Human studies report wrist speeds of 10--53.3 RPM~\cite{vaisman_comparative_2013}. Conscious and involuntary reflexes span 50--100 ms, giving bandwidth frequencies of 10--20 Hz~\cite{forgaard_voluntary_2015}. Minimum wrist angular precision is 3.47\textdegree~\cite{holman_accuracy_2020}.

ADL studies show that $\pm40^\circ$ F/E and $[-10^\circ, 30^\circ]$ R/U allow completion of 22 of the 24 tasks~\cite{ryu_functional_1991}. We align \textit{DexWrist} joint coordinates such that the pitch--yaw output DOFs correspond to F/E and R/U.

\subsubsection{Size and Weight}
NASA anthropometric data give a 95th-percentile male wrist width and height of 61.4~mm, and a forearm length of 349~mm~\cite{anthropometry}, which we shorten by $2.5 \times 61.4$~mm to 195.5~mm to leave ample room for an elbow joint and a P/S DOF upstream, as most robot arms include this DOF. As a drop-in module, we target the $\sim$1~kg weight of common grippers such as the Robotiq 2F series~\cite{robotiq_2f}.

\vspace{-1.5mm}
\subsection{Wrist Design}
\label{sec:wristdesign}

\textit{DexWrist} is a 2-DOF robotic wrist meeting the requirements of Sec. \ref{sec:frs}. Two custom QDD actuators in the forearm each drive one leg of a decoupled 2-(R,RR) parallel kinematic mechanism (PKM), whose output axes intersect at a fixed wrist center with output angles $q_1$ (yaw), $q_2$ (pitch). Roll is assumed upstream, but can be easily added with our modular actuators.

\subsubsection{2-(R, RR) Decoupled Parallel Kinematic Mechanism}
\label{sec:pkm}

\paragraph{Architecture and motion}
We implement a 2-(R,RR) spherical parallel mechanism similar to the Agile Eye~\cite{caron1997agileeye2dof}, where "spherical" denotes pure rotation about the fixed wrist center. Both actuators stay in the forearm rather than rotating with the output flange, reducing distal mass versus U-joint layouts and raising bandwidth. The wrist center is 48.75~mm from the output and 70~mm from the Sec.~\ref{sec:frs} load point.

\begin{figure}[t]
    \centering
    \includegraphics[width=0.9\linewidth]{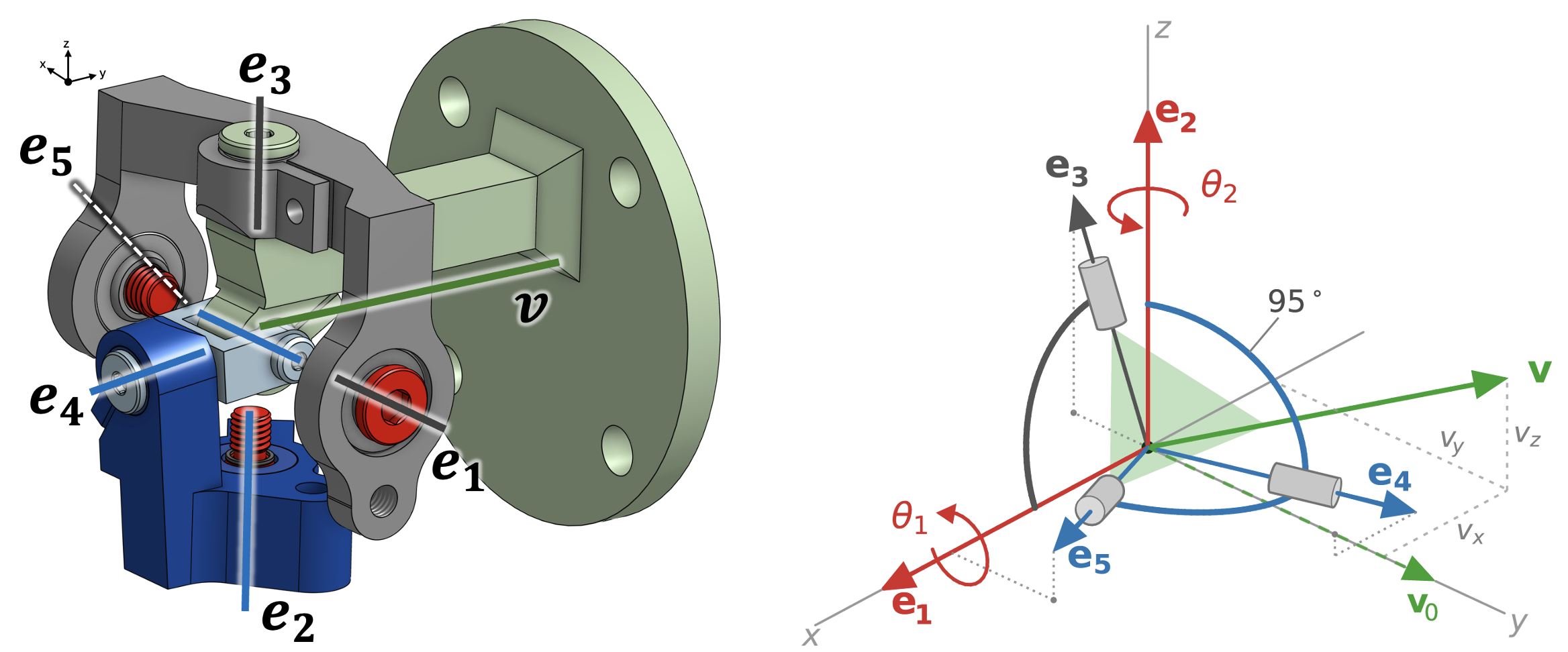}
    \caption{The 2-(R,\,RR) decoupled PKM. \textbf{Left:} Embodiment. The actuated axes $\mathbf{e}_1,\mathbf{e}_2$ (red) are forearm-fixed; the yaw (R, gray) and pitch (RR, blue) legs connect them to the output link (green). All joint axes intersect at the wrist center. \textbf{Right:} Kinematic schematic in the base frame ($x\!\parallel\!\mathbf{e}_1$, $z\!\parallel\!\mathbf{e}_2$, $y\!\parallel\!\mathbf{v}_0$), deflected at $(\theta_1,\theta_2)=(22^\circ,25^\circ)$. Gray cylinders mark the passive bearings on $\mathbf{e}_3,\mathbf{e}_4,\mathbf{e}_5$; $\mathbf{v}_0$ (dashed) is the home output axis. The shaded triangle is the rigid output link, with $\mathbf{e}_3$, $\mathbf{e}_5$, and $\mathbf{v}$ mutually fixed. Arcs denote $90^\circ$ except the $95^\circ$ between $\mathbf{e}_2$ and $\mathbf{e}_4$.}
    \label{fig:pkm}
    \vspace{-2.5mm}
\end{figure}

\paragraph{Geometry selection and structural checks}
We selected link lengths and joint offsets to satisfy the required ROM without self-collision inside the size envelope, and checked strength under four load cases: 5~kg applied along each base axis (X, Y, Z independently) and simultaneous 3~Nm actuation about both DOFs (Fig.~\ref{fig:pkm}).

\paragraph{Forward and inverse kinematics}
The joint axes $\mathbf{e}_1,\dots,\mathbf{e}_5$ and output axis $\mathbf{v}$ are defined in Fig.~\ref{fig:pkm}, in a base frame with $\mathbf{e}_1=\hat{\mathbf{x}}$ (yaw axis), $\mathbf{e}_2=\hat{\mathbf{z}}$ (pitch axis), and home output $\mathbf{v}_0=\hat{\mathbf{y}}$. The yaw leg rotates $\mathbf{e}_3$ about $\mathbf{e}_1$ by $\theta_1$, and the pitch leg rotates $\mathbf{e}_4$ about $\mathbf{e}_2$ by $\theta_2$. All link twists are $90^\circ$ except $\alpha_2=\angle(\mathbf{e}_2,\mathbf{e}_4)=95^\circ$; this $5^\circ$ offset, $\alpha\triangleq\alpha_2-90^\circ$, provides clearance for the yaw axis to reach the full $\pm40^\circ$.

The output orientation follows from the two loop-closure constraints $\mathbf{e}_3^\top\mathbf{v}=0$ and $\mathbf{e}_4^\top\mathbf{e}_5=0$. Writing the output as $R=R_x(q_1)R_z(q_2)$, the yaw constraint is satisfied directly by $q_1=\theta_1$, and the pitch constraint yields the closed form (abbreviating $c_x\!=\!\cos x$, $s_x\!=\!\sin x$):
\begin{equation}
q_2=\operatorname{atan2}\!\left(c_\alpha s_{\theta_2},\ c_\alpha c_{\theta_1} c_{\theta_2}-s_\alpha s_{\theta_1}\right),
\label{eq:fk}
\end{equation}
with output axis $\mathbf{v}=(-\sin q_2,\ \cos q_2\cos q_1,\ \cos q_2\sin q_1)$. In the orthogonal limit $\alpha\!\to\!0$ this reduces to $\tan q_2=\tan\theta_2/\cos q_1$. Inverting, $\theta_1=q_1$ and $\theta_2$ solves $A\cos\theta_2+B\sin\theta_2=C$ with $A=-\tan q_2\cos\alpha\cos q_1$, $B=\cos\alpha$, $C=-\tan q_2\sin\alpha\sin q_1$, taking the root in $|\theta_i|<\tfrac{\pi}{2}$. The complete derivation is on the project website, and the mapping was verified in CAD.

\paragraph{Decoupling and diagonal constraint Jacobian}

We use the standard parallel-robot differential kinematics form $K(\cdot)\,\dot{\boldsymbol{\theta}}=J_0(\cdot)\,\boldsymbol{\omega}$, where $\dot{\boldsymbol{\theta}}=[\dot{\theta}_1\ \dot{\theta}_2]^\top$ are the actuated joint angle rates and $\boldsymbol{\omega}$ is the output flange's angular velocity. $\dot{\theta}_1$ actuates the yaw leg (R) via $\mathbf{e}_1$, and $\dot{\theta}_2$ the pitch leg (RR) via $\mathbf{e}_2$.
The mechanism imposes two constant-angle constraints:
$\mathbf{e}_3^\top \mathbf{v}=\cos\alpha_3$ and $\mathbf{e}_4^\top \mathbf{e}_5=\cos\alpha_4$, with $\alpha_3,\alpha_4$ fixed by the wrist geometry.
Here $\alpha_3=\alpha_4=\pi/2$, so $\mathbf{e}_3^\top \mathbf{v}=0$ and $\mathbf{e}_4^\top \mathbf{e}_5=0$.
Differentiating and using
$\dot{\mathbf{e}}_3=\dot{\theta}_1(\mathbf{e}_1\times\mathbf{e}_3)$,
$\dot{\mathbf{e}}_4=\dot{\theta}_2(\mathbf{e}_2\times\mathbf{e}_4)$,
$\dot{\mathbf{v}}=\boldsymbol{\omega}\times\mathbf{v}$, and
$\dot{\mathbf{e}}_5=\boldsymbol{\omega}\times\mathbf{e}_5$
yields:
\[
K\,\dot{\boldsymbol{\theta}}=J_0\,\boldsymbol{\omega},\quad
K=
\begin{bmatrix}
\mathbf{v}^\top(\mathbf{e}_1\times\mathbf{e}_3) & 0\\
0 & \mathbf{e}_5^\top(\mathbf{e}_2\times\mathbf{e}_4)
\end{bmatrix}
\]
Since $K$ is diagonal, each constraint equation depends on only one actuated rate~\cite{caron1997agileeye2dof}: the velocity constraint Jacobian is decoupled. This is the one-to-one motor-to-DOF mapping in the velocity-constraint sense: each actuator appears in exactly one constraint, paired with one output-link axis, in contrast to coupled parallel wrists, whose non-diagonal constraint Jacobian requires coordinated multi-motor actuation. In projective coordinates the displacement map of this mechanism family is fully decoupled~\cite{kuo2014agileeye}, though this does not extend to the velocity or torque level. The torque-level consequences are derived next.
Within the restricted configuration space $\mathcal{Q}' = \{ (\theta_1,\theta_2): |\theta_i| < \frac{\pi}{2} \}$ set by the mechanical joint limits, $\mathbf{e}_1\times\mathbf{e}_3$ is nonzero by construction and cannot be orthogonal to $\mathbf{v}$, and the rotational constraint on $\theta_2$ keeps $\mathbf{e}_5^\top(\mathbf{e}_2\times\mathbf{e}_4)\neq 0$, so $\det(K)=K_{11}K_{22}\neq 0$ on $\mathcal{Q}'$.
Over the mechanical range $|\theta_i|\le40^\circ$ the mechanism also stays far from degeneracy: $\partial q_2/\partial\theta_2 \ge 0.98$ and the output Jacobian's condition number never exceeds 1.7. The same bound gives existence and uniqueness of the inverse-kinematics root, since $q_2$ is strictly monotone in $\theta_2$.

\paragraph{Statics and torque attribution}
By virtual work, an external moment $\mathbf{m}$ reflects onto the actuators as $\boldsymbol{\tau}=J^{\top}\mathbf{m}$, where $J$ is the output Jacobian ($\boldsymbol{\omega}=J\dot{\boldsymbol{\theta}}$, with $J_0 J=K$). Because $q_1=\theta_1$ exactly while $q_2$ depends on both actuator angles, $J^{\top}$ is upper triangular in $(M_1, M_2)$:
\begin{equation}
\tau_1 = M_1 + M_2\,\frac{\partial q_2}{\partial\theta_1},
\qquad
\tau_2 = M_2\,\frac{\partial q_2}{\partial\theta_2},
\label{eq:statics}
\end{equation}
where $M_1=\mathbf{m}\cdot\mathbf{e}_1$ and $M_2=\mathbf{m}\cdot\mathbf{u}$ are the moment components about the yaw and pitch output axes ($\mathbf{u}=R_x(\theta_1)\hat{\mathbf{z}}=\mathbf{e}_3$). The remaining component is reacted by the structure. A yaw moment never reaches the pitch actuator at any configuration, while a pitch moment reflects onto the yaw actuator in the ratio $(\partial q_2/\partial\theta_1)/(\partial q_2/\partial\theta_2)$: zero on $\theta_2=0$, at most $6\%$ along $\theta_1=0$ where only the $5^\circ$ twist contributes, growing from $15\%$ at $(20^\circ,20^\circ)$ to $31\%$ at $(30^\circ,30^\circ)$ and $50\%$ at the $(40^\circ,40^\circ)$ motor corner. Setting $\alpha=0$ still leaves $25\%$ at $(30^\circ,30^\circ)$, so the coupling is driven by the yaw deflection, not the twist. As a closed form of the kinematics, \eqref{eq:statics} can be compensated exactly. We verified \eqref{eq:statics} on a bench rig by quasi-statically loading each output axis in turn with known masses and reading the opposite actuator over a 2-D pose grid: the yaw-loaded pitch torque is zero as predicted ($2.2$ percentage points RMS, at the fixture noise floor), the pitch-loaded yaw torque matches the predicted coupling in sign and $\theta_2$-parity at every pose ($0.96\times$ aggregate), and the worst residual is $0.044$~N$\cdot$m, $1.2\%$ of rated torque and $7.5\times$ below the backdrive friction that remains the practical limit on torque attribution (protocol and data on the website).

\subsubsection{Quasi-Direct-Drive Actuator Modules}

Quasi-direct-drive (QDD) actuation~\cite{cheetah_actuators} meets our torque, speed, backdrive-torque, and bandwidth requirements: a high-torque-density motor with a low-ratio transmission reduces friction/hysteresis and reflected motor inertia, enabling high-bandwidth torque control and far lower commanded stiffness through the PD gains than in the high-reduction harmonic-drive wrists of commodity arms (e.g., Franka and UR).

\begin{figure}[t!]
    \centering
    \includegraphics[width=0.9\linewidth]{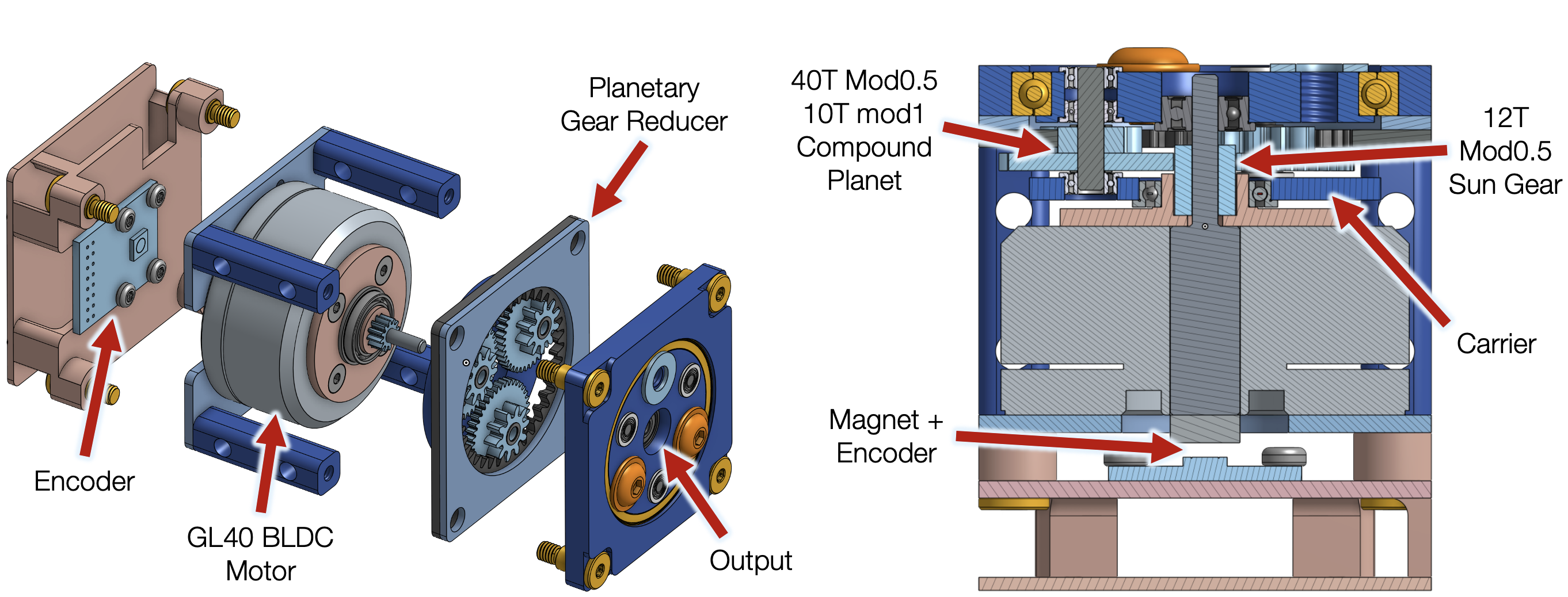}
    \caption{Our custom 13:1 quasi-direct-drive (QDD) actuator.}
    \label{fig:gearbox}
    \vspace{-0.75\baselineskip}
\end{figure}

The actuators use a CubeMars GL40 brushless DC gimbal motor paired with a custom single-stage compound planetary gearbox (Fig.~\ref{fig:gearbox}), since commercial gearboxes did not meet our packaging requirements. A low ratio of 13:1 multiplies the motor rated torque ($\sim$0.25~Nm) past the 3~Nm-class joint target. Reflected rotor inertia is $J_{\mathrm{ref}} = N^2 J_m$, and using GL40 inertia ($J_m=74~\mathrm{g\,cm^2}$) and a worst-case payload inertia ($J_L \approx m r^2$ with $m=5$~kg at $r=70$~mm), we obtain $J_{\mathrm{ref}}/J_L \approx 5\times10^{-2} \ll 1$, consistent with QDD-style low-impedance actuation \cite{cheetah_actuators}. Gear tooth bending stress is checked with the Lewis equation (including dynamic factors) using 1045-steel gears, targeting $\ge 3$ factor of safety.

An AS5047P encoder measures motor position; a moteus-n1 controller, mounted in the actuator and daisy-chained over CAN at 16~V, runs a 1~kHz torque/current loop.
\vspace{-2mm}

\subsection{Mechanical Validation}
Experiments were performed to validate the functional requirements in Table \ref{tab:functional_reqs}. Results are in this table as well.

\subsubsection{Output Torque, Bandwidth, and Backdrive Torque} We measured torque with a Vernier Go force sensor at 70~mm from the wrist center, past the PKM rather than at the actuator: continuous 60~s measurements gave 3.75~Nm on both axes. Rigidly fixing the flange and converting the 10--90\% torque rise time by the first-order relation $B = 0.35/t_r$ gives 10.15~Hz. Backdrive torque with PD gains set to 0 and no feedforward torque is 0.33~Nm.

\subsubsection{Structural Load Capacity} Rigidly mounted to an 8020 testbed, we loaded the output flange (48.75~mm from the wrist center) directly with a crane scale and ratchet straps: axial loading reached 100~kg in both directions, and converting output force to torque about the wrist center gave 14~Nm in both directions about both axes at the hardstops. These are tested-to values, not failure limits.

\subsubsection{Speed, Angular Precision, and Range of Motion} Tracking the flange from center to both endstops with Vernier Video Analysis gave a maximum of 96.6~RPM per axis and a 1.65$^\circ$ SD of start/end position differences over 10 trials. ROM was $[-40^\circ,40^\circ]$ in pitch--yaw.

\subsubsection{Size and Weight} \textit{DexWrist}'s length is 178.2 mm, meeting the requirement. Width and height (64~mm and 66.5~mm, set by the driving links) exceed the 61.4~mm target by only 4\% and 8\%, respectively. The assembly weighs 0.97 kg, below our 1.0 kg target mass.


\section{System-Level Evaluation}
\label{sec:experimentalsetup}
We evaluate how \textit{DexWrist}'s mechanical properties (Sec. \ref{sec:hardware}) translate to system-level gains through three experiments: a simulated workspace comparison, a teleoperated user study, and autonomous task completion on two representative tasks. The latter two use behavior cloning (BC). A human teleoperates the robot to perform a task, and each successful example becomes a \emph{demonstration}: per-timestep \emph{observations} (RGB image and proprioceptive state) paired with the operator's commanded \emph{actions}. From a dataset of demonstrations we train a \emph{policy} $\pi_\theta(a_{t:t+c} \mid o_{t-k:t})$ via supervised learning to model a distribution over action chunks given a history of observations, with performance measured by success rate and task completion time.

\vspace{-1mm}
\subsection{System Integration}
\label{sec:integration}
We tested \textit{DexWrist} on two primary platforms, each stressing one design property: (i) the AgileX PiPER, whose spaced-apart serial wrist joints limit constrained-space kinematics (testing the PKM), and (ii) the UR3e, whose stiff harmonic drives limit contact-rich manipulation (testing the QDD actuation). For the wiping task we additionally test on the Franka FR3, whose joint-torque sensing supports high-quality joint-impedance control, to isolate the wrist's contribution from the low-level controller. All platforms use the AgileX PiPER parallel-jaw (ALOHA-style) gripper.

\subsubsection{Manipulation in Constrained Spaces}
We study how \textit{DexWrist}'s co-located DOFs and small size affect manipulation in constrained spaces.
\begin{compactitem}
    \item \textbf{AgileX \!+\! stock wrist}: Uses backdrivable QDD actuators for contact-rich tasks, but its wrist joints are not co-located, complicating reorientation in tight spaces.
    \item \textbf{AgileX \!+\! DexWrist}: We replace the AgileX's last two joints with \textit{DexWrist}, maintaining 6-DOF.
\end{compactitem}

\subsubsection{Dynamic and Contact-Rich Manipulation}
We study how \textit{DexWrist}'s QDD actuation and one-to-one mapping affect proprioception, speed, and simple compliant control.
\begin{compactitem}
    \item \textbf{UR3e \!+\! stock wrist}: Uses stiff, high-reduction harmonic drives that fault the robot upon rigid contact in position control. A closed-loop Cartesian admittance controller~\cite{ott2010unified} allows contact by adjusting motor commands from sensed force, but restricts us to EE pose commands, as the UR3e's only F-T sensor is at the EE.
    \item \textbf{UR3e \!+\! DexWrist}: We lock the UR3e's Wrist1 and Wrist2 joints and attach \textit{DexWrist} at the end of the chain to maintain 6-DOF. Its backdrivable, low-inertia actuation supports joint position control without faulting the robot, while allowing stiffness modulation.
    \item \textbf{Franka \!+\! stock wrist}: A torque-controlled arm under a joint-impedance controller, among the best general-purpose compliant control on commercial arms.
    \item \textbf{Franka \!+\! DexWrist}: We lock the Franka's J5 and J6 joints and attach \textit{DexWrist}, keeping the same joint-impedance controller and joint-space action so only the wrist hardware differs between the conditions.
\end{compactitem}

Observation and teleoperation interfaces for demonstration collection and policy learning are summarized in Table~\ref{tab:obs_actions}. The UR3e's leader arm is a second UR3e, and the Franka's is a GELLO~\cite{wu2024gello}. We keep end-effector actions on the AgileX and joint-position actions on the UR3e + \textit{DexWrist} to test whether \textit{DexWrist}'s benefits hold across both action representations. On the stock UR3e, the Cartesian admittance controller needed for wiping requires EE actions. AgileX policies run at 30~Hz (200~Hz low-level) and UR3e policies at 50~Hz (500~Hz). Inverse kinematics (IK) converts between EE pose and joint space as needed.

\begin{table}[t]
  \centering
  \setlength{\tabcolsep}{5pt}
  \renewcommand{\arraystretch}{1.1}
  \caption{Sensing and action interfaces for each setup (EE = end-effector). Each column covers the arm's stock-wrist and \textit{DexWrist}; where they differ, cells list stock~/~\textit{DexWrist}.}
  \label{tab:obs_actions}
  \footnotesize
  \begin{tabular}{llll}
    \toprule
      & AgileX
      & UR3e
      & Franka \\
    \midrule
    \textbf{Controller}
      & SpaceMouse
      & Lead--follow
      & Lead--follow \\
    \addlinespace[0.2em]
    \textbf{Camera}
      & Intel D405
      & OAK-1 W
      & RealSense D435 \\
    \addlinespace[0.2em]
    \textbf{Obs}
      & \begin{tabular}[t]{@{}l@{}}Wrist RGB\\+ EE Pose\end{tabular}
      & \begin{tabular}[t]{@{}l@{}}Overhead RGB\\+ Joint Pos.\end{tabular}
      & \begin{tabular}[t]{@{}l@{}}RGB\\+ Joint Pos.\end{tabular} \\
    \addlinespace[0.2em]
    \textbf{Action}
      & EE Pose
      & \begin{tabular}[t]{@{}l@{}}EE Pose~/\\Joint Pos.\end{tabular}
      & Joint Pos. \\
    \bottomrule
  \end{tabular}
\end{table}

\begin{figure}[t!]
    \centering
    \includegraphics[width=1.0\linewidth]{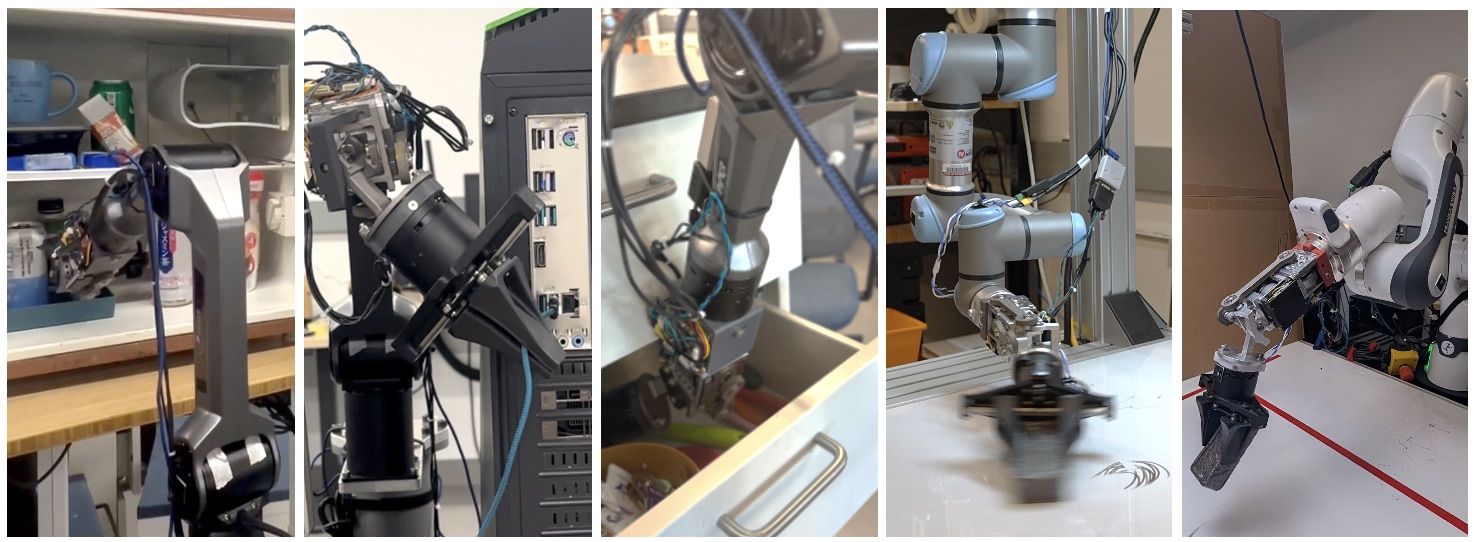}
    \caption{Teleoperation tasks, from left to right: picking from a cluttered fridge, cable unplugging, picking from a deep drawer, and wiping a whiteboard on the UR3e and on the Franka. The fridge and wiping tasks also serve as the BC testbeds (constrained and dynamic+contact-rich, respectively).}
    \label{fig:tasks}
    \vspace{-2.5mm}
\end{figure}

\vspace{-1.5mm}
\subsection{Experiments}

\subsubsection{Workspace Comparison}
We quantify the workspace benefit of \textit{DexWrist}'s co-located pitch--yaw center by comparing the AgileX PiPER with its stock wrist and with \textit{DexWrist} in a simulated confined-space task. In a PyBullet environment with a deep cabinet model, we sample target points uniformly inside the volume and solve IK for a fixed end-effector orientation from a fixed nominal start configuration. A point is reachable if a collision-free IK solution exists at the goal pose (checking self- and cabinet collisions). We do not collision-check a full motion trajectory.

\subsubsection{Teleoperated Task User Study}
\label{sec:teleopexperiments}
Collecting numerous high-quality demonstrations quickly is critical to training high-performing policies. Under BC, errors accumulate as $\mathcal{O}(\epsilon T^2)$ (worst case) for trajectories with $T$ time steps~\cite{ross2011reduction}, where $\epsilon$ is the per-step error rate, so shorter demonstrations tighten the worst-case compounding bound. Through a user study covering four real-world, confined and contact-rich tasks (Fig.~\ref{fig:tasks}), we record task completion time, environment resets, and total operator time, collecting at least 35 demos per setup (100+ for the two policy-learning tasks), more than 500 in total. For the wiping task, we record the same measurements on both Franka conditions under a strong impedance-control baseline.

\begin{compactitem}
    \item \textbf{Picking From a Cluttered Fridge (AgileX)}: Pick a highly occluded can without disturbing its surroundings.
    \item \textbf{Whiteboard Wiping (UR3e, Franka)}: Wipe scribbles and dots off a whiteboard.
    \item \textbf{Cable Unplugging (AgileX)}: Unplug a cable through the narrow gap between a monitor and a computer.
    \item \textbf{Picking From a Drawer (AgileX)}: Pick up a cup from deep inside a drawer. 
\end{compactitem}

\subsubsection{Autonomous Task Completion using Behavior Cloning}
\label{sec:policybackground}
We train diffusion policies~\cite{chi2024diffusionpolicy} on the fridge task (constrained, AgileX) and the wiping task (contact-rich, UR3e) to test whether \textit{DexWrist}'s design benefits translate to gains in real-world autonomous performance. For each of the four configurations, we evaluated policies at six training checkpoints (epochs 75, 150, 225, 300, 375, 750). Success rates plateaued across these checkpoints rather than degrading. Because offline validation loss poorly predicts closed-loop success, we select and report the best-performing checkpoint by closed-loop rollout, following standard practice~\cite{mandlekar2021robomimic,chi2024diffusionpolicy}. The same checkpoint grid and selection rule apply to every configuration. Initial object poses and task parameters (e.g., scribble location and size, object placement) are randomized every reset. We formally define the autonomous tasks below.

\begin{figure}[t]
    \centering
    \includegraphics[width=0.8\linewidth]{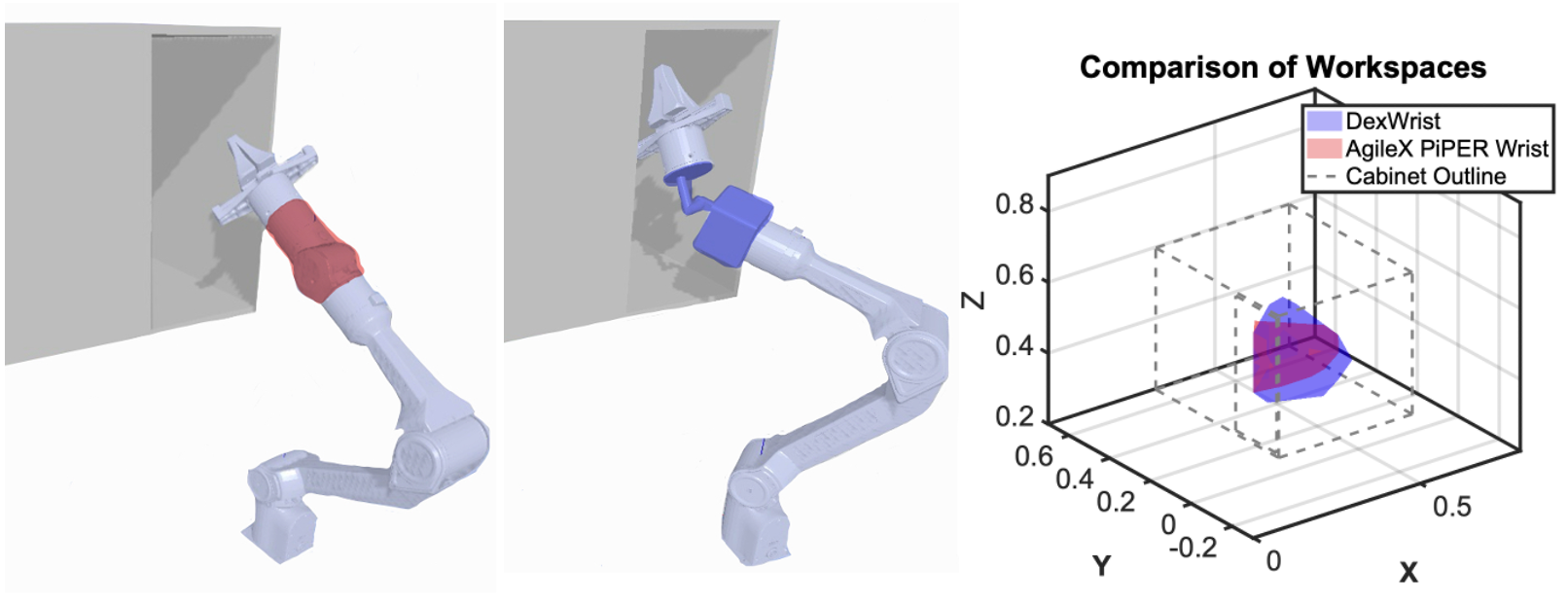}
    \caption{\textbf{Left:} Serial wrist baseline (AgileX). \textbf{Middle:} \textit{DexWrist}. \textbf{Right:} Workspace comparison by reaching through the narrow opening of a kitchen cabinet, with \textit{DexWrist} achieving an 88\% improvement over the AgileX.}
    \label{fig:kinematics}
    \vspace{-2.5mm}
\end{figure}

\paragraph{\textbf{Constrained Pick-and-Place (AgileX)}}
\label{constrained_pick_and_place}
The task is to retrieve an occluded, flattened soda can from deep within a cluttered refrigerator and place it on the table. Failure occurs if any object is knocked over, the camera disconnects, or the refrigerator is displaced. The can sits directly beneath a shelf, so there is no top-down access: the end-effector must bend sharply and compactly to one side to reach it. Object poses are randomized every reset, with greater variation for the target.

We use the same control method on both robots: each policy predicts a chunk of 16 consecutive actions $\hat{\mathbf{a}}_{t:t+15}$, where each 10D action $\hat{\mathbf{a}}_k = [{}^{B} \mathbf{r}_k^{\textit{TCP}} ; {}^B\mathcal{R}_k^{\textit{TCP}} ; g_k] \in \mathbb{R}^3 \times \mathbb{R}^6 \times \{-1, 0, 1\}$ stacks a target end-effector position in the robot base frame, a continuous 6D rotation~\cite{DBLP:journals/corr/abs-1812-07035}, and a discrete gripper command. The first 8 actions are executed before re-planning; pose targets stream at 30\,Hz to a 200\,Hz low-level joint controller.

\paragraph{\textbf{Dynamic Wiping}}
The robot wipes scribbles of various sizes and shapes, and dots 2--6 cm in diameter, drawn at random locations on a whiteboard. Success is erasing more than 50\% of a scribble. A failure occurs if an emergency stop triggers. We run the same task on the Franka FR3 to compare \textit{DexWrist} against a strong compliant-control baseline.

The stock-wrist policy predicts end-effector pose targets tracked by the Cartesian admittance controller (Sec.~\ref{sec:integration}); the \textit{DexWrist} policy outputs joint position targets (Table~\ref{tab:obs_actions}). Both policies condition on 2 steps of history and predict a 96-action horizon, executing the first 32 before re-planning.

\section{Results}
\label{results}

\subsection{Workspace Comparison}
Collision-free reachable targets increase by 88\% with \textit{DexWrist} relative to the stock AgileX wrist (Fig.~\ref{fig:kinematics}). The stock wrist fails primarily at targets deep in the cabinet and near its walls, where the serial joint layout forces the arm into self- or cabinet collisions to reach the required orientation.

\vspace{-2mm}
\subsection{Teleoperation User Study Results}
\begin{figure}
    \centering
    \includegraphics[width=1.0\linewidth]{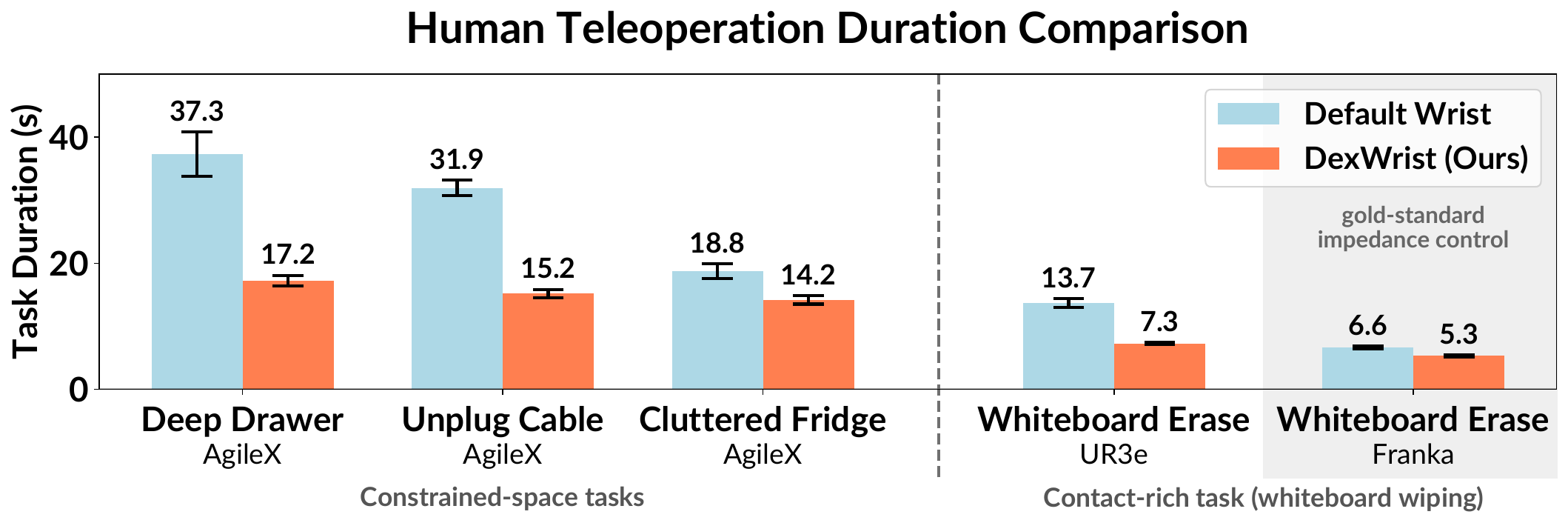}
    \caption{Durations of successful demonstrations. \textit{DexWrist} reduces durations across constrained and dynamic tasks; the two whiteboard-erase groups compare the same task on the stiff UR3e and the torque-controlled Franka. Error bars: SEM. Lower is better.}
    \label{fig:trajComparisons}
    \vspace{-2mm}
\end{figure}

As shown in Fig.~\ref{fig:trajComparisons}, \textit{DexWrist} reduced the average duration of successful trajectories by 1.3--2.2$\times$ relative to the stock AgileX and UR3e arms. It also cut collection effort on those tasks, requiring 39.0--69.3\% less operator time and fewer resets (Table~\ref{tab:teleop_table}). On the Franka wiping comparison, where the strong baseline leaves less to improve, the reductions are smaller, with demonstrations 1.2$\times$ shorter.

\begin{table}
  \centering
   \caption{\textit{DexWrist} decreases total operator time (including resets and failed attempts) and the resets needed to obtain successful demonstrations. Resets follow a severe collision, a knocked-over object, or a failed grasp.}
   \label{tab:teleop_table}
  \footnotesize
  \setlength{\tabcolsep}{4pt}
  \renewcommand{\arraystretch}{0.95}
  \begin{tabular}{ll|cc}
    \toprule
    \textbf{Metric (Mean)} & \textbf{Task} & \textbf{Base Robot} & \textbf{w/ \textit{DexWrist} (ours)} \\
    \midrule
    \multirow{5}{*}{Operator Time (s)} & Fridge & 63.5 & \textbf{38.7} \\
                                & Cable & 76.3 & \textbf{28.0} \\
                                & Drawer & 57.2 & \textbf{29.6} \\
                                & Wipe (UR3e) & 21.5 & \textbf{6.6} \\
    \cmidrule(lr){2-4}
                                & Wipe (Franka) & 24.2 & \textbf{22.4} \\
                              
    \midrule
    \multirow{5}{*}{Operator Resets} & Fridge & 1.7 & \textbf{1.0} \\
                                & Cable & 0.6 & \textbf{0.4} \\
                                & Drawer & 0.6 & \textbf{0.3} \\
                                & Wipe (UR3e) & 0.2 & \textbf{0.0} \\
    \cmidrule(lr){2-4}
                                & Wipe (Franka) & 0.2 & \textbf{0.0} \\
    \bottomrule
  \end{tabular}
\end{table}

\vspace{-2mm}
\subsection{Autonomous Task Results}
\label{sec:policylearning}
\textbf{Constrained Pick-and-Place (AgileX):} The best-performing policy trained for AgileX + \textit{DexWrist} achieved a 50\% relative improvement in success rate over the policy trained for the default AgileX. Qualitatively, the policy trained for the stock wrist was observed to violently disrupt the environment more frequently (e.g., displacing the fridge, knocking shelves out). On average, the AgileX + \textit{DexWrist} completed the task \textbf{3.24$\times$} faster than the stock system.

\begin{figure}[t]
    \vspace{-0.5mm}
    \centering
    \includegraphics[width=1.0\linewidth]{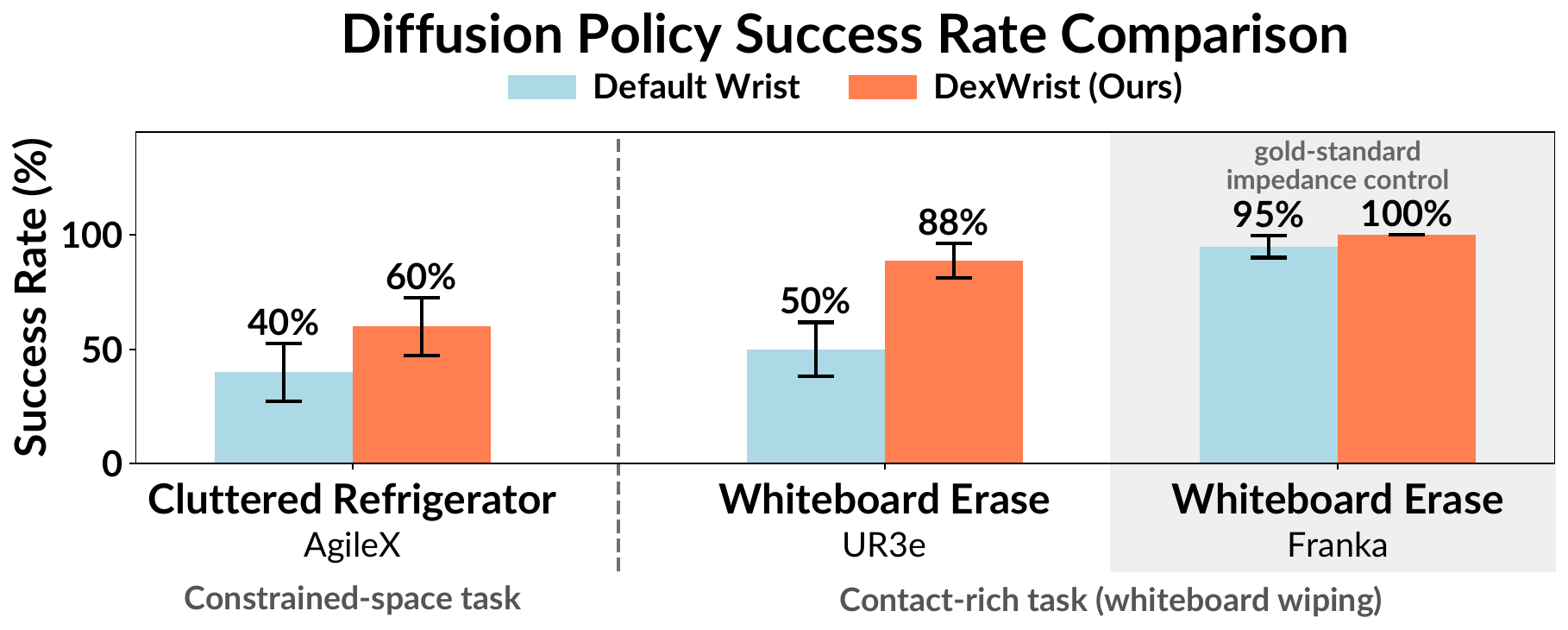}
    \caption{Comparison of policy success rates across baseline arms and their \textit{DexWrist} versions. \textit{DexWrist} substantially improves success on the UR3e and AgileX; on the Franka, whose strong joint-impedance baseline already succeeds, it matches the baseline. Error bars: SEM, with at least 15 trials each (Table~\ref{tab:policy-comparison}). Higher is better.}
    \label{fig:policy_success}
    \vspace{-2mm}
\end{figure}

\textbf{Dynamic Wiping (UR3e, Franka):} We evaluate wiping on two arms at opposite ends of the actuation spectrum. On the stiff, harmonic-drive UR3e, the UR3e + \textit{DexWrist} achieved a 76\% relative improvement in success rate over the stock UR3e (Fig.~\ref{fig:policy_success}) and completed the task \textbf{4.92$\times$} faster (Table~\ref{tab:policy-comparison}); the baseline often failed to erase even half of a scribble, while the UR3e + \textit{DexWrist} consistently erased nearly all of it. To confirm this gap is not an artifact of the stock UR3e's admittance controller, a separate teleoperated study on a dot-wiping variant ($n{=}42$ trials per condition) verified that our controller performs comparably to UR's native force-mode controller (completion time $p=0.70$, success $p=0.78$; details on the project website). On the Franka FR3, whose high-quality joint-impedance controller already succeeds, adding \textit{DexWrist} left success statistically unchanged (stock 19/20, \textit{DexWrist} 20/20); with success saturated for both conditions, the informative metric is completion time, where \textit{DexWrist} was \textbf{1.43$\times$} faster with the same controller and action space (one-sided Mann--Whitney $p\approx9\times10^{-5}$).

\begin{table}[t!]
  \centering
  \caption{Autonomous task completion time statistics for successful trials at each system's best checkpoint (at least $N=15$ per system). \textit{DexWrist} substantially increases completion speed.}
  \label{tab:policy-comparison}
  \begin{tabular}{l|ccc}
    \toprule
    \multirow{2}{*}{\textbf{System}} & \multicolumn{3}{c}{\textbf{Policy Task Completion Time (s)}} \\
    \cmidrule(lr){2-4}
    & Mean $\pm$~SD & Min & Max \\
    \midrule
    AgileX \!+\! stock wrist & 91.0 $\pm$ \; 7.9 & 55.2 & 134.2 \\
    AgileX \!+\! \textbf{DexWrist (Ours)} & \textbf{28.1} $\pm$ \; \textbf{2.2} & \textbf{20.5} & \textbf{49.0} \\
    UR3e \!+\! stock wrist & 21.2 $\pm$ \; 10.5 & 12.7 & 40.6 \\
    UR3e \!+\! \textbf{DexWrist (Ours)} & \textbf{4.3} $\pm$ \; \textbf{1.2} & \textbf{1.9} & \textbf{6.5} \\
    \midrule
    Franka \!+\! stock wrist & 6.1 $\pm$ \; 1.4 & 4.3 & 9.0 \\
    Franka \!+\! \textbf{DexWrist (Ours)} & \textbf{4.3} $\pm$ \; \textbf{1.3} & \textbf{2.9} & \textbf{7.8} \\
    \bottomrule
  \end{tabular}
  \vspace{-3mm}
\end{table}

\section{Discussion}
\label{sec:discussion}

\textit{DexWrist}'s novel combination of quasi-direct-drive actuation and a decoupled parallel kinematic mechanism addresses many shortcomings of current wrist designs. The design was validated in Sec.~\ref{sec:hardware}, and co-location produced an 88\% workspace increase in simulation.

At the system level, these properties translate to more efficient teleoperation. \textit{DexWrist} reduced data collection time by 39.0--69.3\% on the AgileX and UR3e tasks while reducing resets, with 1.3--2.2$\times$ shorter demonstrations. We attribute these results to \textit{DexWrist}'s (1) co-located DOFs reorienting the end-effector by moving only the short distal links, requiring less arm articulation in confined spaces; (2) task workspace staying far from the near-singular configurations where small end-effector adjustments demand large joint velocities; and (3) quasi-direct-drive actuation stabilizing contact without an admittance controller. Lower operator time and reset count yield more data per unit time, and shorter demonstrations tighten the compounding-error bound (Sec.~\ref{sec:teleopexperiments}); both carried over to policy performance.

Under diffusion-policy evaluation, these mechanisms produced the 50--76\% success-rate and 3--5$\times$ completion-time gains on the AgileX and UR3e. In the constrained pick-and-place task, \textit{DexWrist} reorients the gripper inside the fridge by moving only the distal, co-located pitch/yaw links, requiring little motion from the arm. The stock wrist, whose final two DOFs sit further upstream, must instead articulate the whole arm, bringing its links closer to surrounding objects and increasing the chance of collisions. The learned policy imitates these more compact demonstrations and contacts objects less often, consistent with the lower teleoperation reset counts (Table~\ref{tab:teleop_table}). The AgileX's kinematic chain also passes through a near-singular configuration as it enters the fridge, where IK becomes poorly conditioned and unreliable. The autonomous speedup exceeds the 1.3--2.2$\times$ demonstration-length ratio because recovery behavior inflates the baseline's rollout times (Table~\ref{tab:policy-comparison}).

In the wiping task, the stock UR3e's stiff harmonic drives resisted backdriving, producing contact forces beyond what the F-T sensor and admittance controller could absorb and sometimes faulting the robot even on its least restrictive safety settings. At contact, peak force scales with the endpoint mechanical impedance and approach speed ($F \sim Zv$), and the reflected rotor inertia that dominates $Z$ scales as $N^2 J_m$: between our 13:1 reduction and the 80--160:1 of harmonic drives, this factor alone is $40$--$150\times$, so \textit{DexWrist} bounds contact force passively, before any control loop can react. Because the backdrivable joints are PD-controlled, contact also produces a steady-state tracking error $e \approx \tau_{\mathrm{ext}}/K_p$ on the pitch axis (an observable contact signal in the policy's joint-position inputs, where a stiff harmonic joint deflects negligibly), which may have allowed the policy to learn when to wipe. This benefit is largest where the arm's own contact capability is weakest: on the stiff, force-sensing-limited UR3e, \textit{DexWrist} yields large success and speed gains, whereas on the torque-controlled Franka it preserves success and still improves speed with the same controller in both conditions, isolating the improvement to the wrist hardware rather than the control loop. Across both tasks, we hypothesize that shorter demonstrations, compact motions from co-located DOFs, and transparent proprioception from decoupled QDD actuation make expert behavior more learnable through BC.

\section{Limitations and Future Work}
\label{sec:limitations}
Being a parallel wrist, \textit{DexWrist} is mechanically more complex than a serial wrist. Its parallel linkages also introduce some backlash, making it better suited to contact-rich manipulation than high-precision positioning. The decoupled mapping gives no load sharing, so each actuator must be sized for its axis's full torque. The decoupling holds exactly at the velocity-constraint level (Sec.~\ref{sec:pkm}); at the torque level, a pitch moment reflects onto the yaw actuator by the closed-form ratio in \eqref{eq:statics}, reaching $50\%$ at the corners, and the measured unmodelled residual is $1.2\%$ of rated torque under quasi-static loading, leaving dynamic cross-axis coupling to future work. The $\pm40^\circ$ range of motion per axis meets the ADL targets we adopt (Sec.~\ref{sec:frs}) but not the full human wrist flexion-extension range. Width and height exceed the target by 4\% and 8\%, respectively. A cover over the linkages and a cable management solution would add polish to the design.

Future work includes adding the roll (P/S) DOF, switching to frameless BLDC motors for higher torque and lower inertia, exploring reinforcement learning via torque control, and studying human demonstration transfer given \textit{DexWrist}'s kinematic similarity to the human wrist.

\section*{Acknowledgments}
We thank the members of the Improbable AI lab, especially Nolan Fey and Branden Romero, for the helpful discussions and feedback on the paper. We are grateful to MIT Supercloud and the Lincoln Laboratory Supercomputing Center for providing HPC resources. We acknowledge AgileX for providing us a custom version of their PiPER firmware.

\bibliographystyle{IEEEtran}
\bibliography{IEEEabrv,citations}

\begin{thebibliography}{10}
\providecommand{\url}[1]{#1}
\csname url@rmstyle\endcsname
\providecommand{\newblock}{\relax}
\providecommand{\bibinfo}[2]{#2}
\providecommand\BIBentrySTDinterwordspacing{\spaceskip=0pt\relax}
\providecommand\BIBentryALTinterwordstretchfactor{4}
\providecommand\BIBentryALTinterwordspacing{\spaceskip=\fontdimen2\font plus
\BIBentryALTinterwordstretchfactor\fontdimen3\font minus
  \fontdimen4\font\relax}
\providecommand\BIBforeignlanguage[2]{{%
\expandafter\ifx\csname l@#1\endcsname\relax
\typeout{** WARNING: IEEEtran.bst: No hyphenation pattern has been}%
\typeout{** loaded for the language `#1'. Using the pattern for}%
\typeout{** the default language instead.}%
\else
\language=\csname l@#1\endcsname
\fi
#2}}

\bibitem{open_x_embodiment_rt_x_2023}
{OX-Embodiment Collab.}, ``Open {X-Embodiment}: Robotic learning datasets and
  {RT-X} models,'' \emph{arXiv preprint arXiv:2310.08864}, 2023.

\bibitem{noauthor_franka_nodate}
\BIBentryALTinterwordspacing
``Franka {Emika} {Panda} robot - {RoboDK}.'' [Online]. Available:
  \url{https://robodk.com/robot/Franka/Emika-Panda}
\BIBentrySTDinterwordspacing

\bibitem{universal_robots_products}
\BIBentryALTinterwordspacing
{Universal Robots}, ``{Collaborative Robots from Universal Robots}.'' [Online].
  Available: \url{https://www.universal-robots.com/products/}
\BIBentrySTDinterwordspacing

\bibitem{cheetah_actuators}
P.~M. Wensing, \emph{et~al.}, ``Proprioceptive actuator design in the {MIT
  Cheetah}: Impact mitigation and high-bandwidth physical interaction for
  dynamic legged robots,'' \emph{IEEE Trans. Robot.}, vol.~33, no.~3, pp.
  509--522, 2017.

\bibitem{negrello_compact_2019}
F.~Negrello, \emph{et~al.}, ``A {Compact} {Soft} {Articulated} {Parallel}
  {Wrist} for {Grasping} in {Narrow} {Spaces},'' \emph{IEEE Robot. Autom.
  Lett.}, vol.~4, no.~4, pp. 3161--3168, 2019.

\bibitem{bajajparallelwrist}
N.~M. Bajaj and A.~M. Dollar, ``Kinematic optimization of a 2-dof u, 2pss
  parallel wrist device,'' in \emph{Proc. ASME Int. Design Eng. Tech. Conf.
  (IDETC-CIE)}, vol.~5A, 2019, p. V05AT07A027.

\bibitem{petric_hammering_2017}
T.~Petri{\v{c}}, \emph{et~al.}, ``Hammering does not fit {Fitts'} law,''
  \emph{Front. Comput. Neurosci.}, vol.~11, p.~45, 2017.

\bibitem{bajaj_state_2019}
N.~M. Bajaj, \emph{et~al.}, ``State of the {Art} in {Artificial} {Wrists}: {A}
  {Review} of {Prosthetic} and {Robotic} {Wrist} {Design},'' \emph{IEEE Trans.
  Robot.}, vol.~35, no.~1, pp. 261--277, 2019.

\bibitem{noauthor_piper_nodate}
\BIBentryALTinterwordspacing
``\BIBforeignlanguage{en}{{PiPER}}.'' [Online]. Available:
  \url{https://global.agilex.ai/products/piper}
\BIBentrySTDinterwordspacing

\bibitem{unitree_h1_2}
\BIBentryALTinterwordspacing
{Unitree Robotics}, ``Unitree {H1-2} {F}ull-size {U}niversal {H}umanoid
  {R}obot,'' 2024. [Online]. Available: \url{https://www.unitree.com/h1}
\BIBentrySTDinterwordspacing

\bibitem{galaxea_r1_manual}
\BIBentryALTinterwordspacing
{Galaxea Dynamics}, ``Galaxea {R1} {H}ardware {G}uide,'' 2024. [Online].
  Available: \url{https://docs.galaxea-ai.com/Guide/R1/Hardware_Guide/}
\BIBentrySTDinterwordspacing

\bibitem{openarm}
\BIBentryALTinterwordspacing
{Enactic, Inc.}, ``{OpenArm}: A fully open-source humanoid arm for physical
  {AI} research,'' 2024. [Online]. Available: \url{https://openarm.dev/}
\BIBentrySTDinterwordspacing

\bibitem{milazzovariablewrist}
G.~Milazzo, \emph{et~al.}, ``Modeling and control of a novel variable stiffness
  three dofs wrist,'' \emph{Int. J. Robot. Res.}, vol.~43, no.~12, pp.
  1898--1915, 2024.

\bibitem{sunmodularwrist}
H.~Sun, \emph{et~al.}, ``Compact modular robotic wrist with variable stiffness
  capability,'' \emph{IEEE Trans. Robot.}, vol.~41, pp. 141--158, 2025.

\bibitem{BaggetaCableWrist}
M.~Baggetta, \emph{et~al.}, ``Virtual and physical prototyping of a
  cable-driven compliant robotic wrist,'' \emph{IEEE/ASME Trans. Mechatron.},
  vol.~30, no.~5, pp. 4000--4010, 2025.

\bibitem{biflex}
G.-C. Jeong, \emph{et~al.}, ``{BiFlex}: A passive bimodal stiffness flexible
  wrist for manipulation in unstructured environments,'' \emph{IEEE Robot.
  Autom. Lett.}, vol.~10, no.~10, pp. 10\,783--10\,790, 2025.

\bibitem{sofka_omni-wrist_2006}
J.~Sofka, \emph{et~al.}, ``Omni-{Wrist} {III} - a new generation of pointing
  devices. {Part} {II}. {Gimbals} systems - control,'' \emph{IEEE Trans.
  Aerosp. Electron. Syst.}, vol.~42, no.~2, pp. 726--734, 2006.

\bibitem{alvarez_design_2019}
J.~B.~R. Alvarez, \emph{et~al.}, ``Design and {Development} of a {Carpal}
  {Wrist} {Robotic} {Manipulator},'' in \emph{Proc. IEEE Int. Conf. Humanoid,
  Nanotechnol., Inf. Technol., Commun. Control, Environ., Manag. (HNICEM)},
  2019, pp. 1--5.

\bibitem{damerla_design_2022}
R.~Damerla, \emph{et~al.}, ``Design and {Testing} of a {Novel},
  {High}-{Performance} {Two} {DoF} {Prosthetic} {Wrist},'' \emph{IEEE Trans.
  Med. Robot. Bionics}, vol.~4, no.~2, pp. 502--519, 2022.

\bibitem{bytewrist}
J.~Tian, \emph{et~al.}, ``{ByteWrist}: A parallel robotic wrist enabling
  flexible and anthropomorphic motion for confined spaces,'' \emph{arXiv
  preprint arXiv:2509.18084}, 2025.

\bibitem{caron1997agileeye2dof}
F.~Caron, ``Analyse et conception d'un manipulateur parall{\`e}le sph{\'e}rique
  {\`a} deux degr{\'e}s de libert{\'e} pour l'orientation d'une cam{\'e}ra,''
  Master's thesis, Laval University, Quebec, Canada, 1997.

\bibitem{damiao_j4310}
\BIBentryALTinterwordspacing
{Shenzhen Damiao Technology Co., Ltd.}, ``{DM-J4310-2EC V1.1 Gear Motor User
  Manual},'' 2023. [Online]. Available:
  \url{https://files.seeedstudio.com/products/Damiao/DM-J4310-en.pdf}
\BIBentrySTDinterwordspacing

\bibitem{pando_characterization_2013}
A.~Pando and S.~Charles, ``Characterization of wrist kinetics during activities
  of daily living,'' \emph{J. Undergrad. Res.}, vol. 2013, no.~1, 2013, art.
  no. 1966.

\bibitem{nasa_std_5017b}
\emph{NASA-STD-5017B: Design and Development Requirements for Mechanisms},
  National Aeronautics and Space Administration (NASA) NASA Standard
  NASA-STD-5017B, 2022.

\bibitem{vaisman_comparative_2013}
L.~Vaisman, \emph{et~al.}, ``A {Comparative} {Analysis} of {Speed} {Profile}
  {Models} for {Wrist} {Pointing} {Movements},'' \emph{IEEE Trans. Neural Syst.
  Rehabil. Eng.}, vol.~21, no.~5, pp. 756--766, 2013.

\bibitem{forgaard_voluntary_2015}
C.~J. Forgaard, \emph{et~al.}, ``Voluntary reaction time and long-latency
  reflex modulation,'' \emph{J. Neurophysiol.}, vol. 114, no.~6, pp.
  3386--3399, 2015.

\bibitem{holman_accuracy_2020}
M.~E. Holman, \emph{et~al.}, ``\BIBforeignlanguage{en}{Accuracy and precision
  of a wrist movement when vibrotactile prompts inform movement speed},''
  \emph{\BIBforeignlanguage{en}{Somatosens. Mot. Res.}}, vol.~37, no.~3, pp.
  165--171, 2020.

\bibitem{ryu_functional_1991}
J.~Ryu, \emph{et~al.}, ``Functional ranges of motion of the wrist joint,''
  \emph{J. Hand Surg.}, vol.~16, no.~3, pp. 409--419, 1991.

\bibitem{anthropometry}
\BIBentryALTinterwordspacing
``{ANTHROPOMETRY} {AND} {BIOMECHANICS}.'' [Online]. Available:
  \url{https://msis.jsc.nasa.gov/sections/section03.htm}
\BIBentrySTDinterwordspacing

\bibitem{robotiq_2f}
\BIBentryALTinterwordspacing
``\BIBforeignlanguage{en}{{2F}-85 and {2F}-140 {Robot} {Grippers} from
  {Robotiq} {\textbar} {Electromate} {Inc}}.'' [Online]. Available:
  \url{https://www.electromate.com/2f-85-and-2f-140-grippers/}
\BIBentrySTDinterwordspacing

\bibitem{kuo2014agileeye}
C.-H. Kuo, \emph{et~al.}, ``A non-overconstrained variant of the {Agile} {Eye}
  with a special decoupled kinematics,'' \emph{Robotica}, vol.~32, pp.
  889--905, 2014.

\bibitem{ott2010unified}
C.~Ott, \emph{et~al.}, ``Unified impedance and admittance control,'' in
  \emph{Proc. IEEE Int. Conf. Robot. Autom. (ICRA)}, 2010, pp. 554--561.

\bibitem{wu2024gello}
P.~Wu, \emph{et~al.}, ``{GELLO}: A general, low-cost, and intuitive
  teleoperation framework for robot manipulators,'' in \emph{Proc. IEEE/RSJ
  Int. Conf. Intell. Robots Syst. (IROS)}, 2024.

\bibitem{ross2011reduction}
S.~Ross, \emph{et~al.}, ``A reduction of imitation learning and structured
  prediction to no-regret online learning,'' in \emph{Proc. Int. Conf. Artif.
  Intell. Stat. (AISTATS)}, 2011, pp. 627--635.

\bibitem{chi2024diffusionpolicy}
C.~Chi, \emph{et~al.}, ``Diffusion policy: Visuomotor policy learning via
  action diffusion,'' \emph{Int. J. Robot. Res.}, vol.~44, no. 10-11, pp.
  1684--1704, 2025.

\bibitem{mandlekar2021robomimic}
A.~Mandlekar, \emph{et~al.}, ``What matters in learning from offline human
  demonstrations for robot manipulation,'' in \emph{Proc. Conf. Robot Learn.
  (CoRL)}, 2021.

\bibitem{DBLP:journals/corr/abs-1812-07035}
Y.~Zhou, \emph{et~al.}, ``On the continuity of rotation representations in
  neural networks,'' \emph{arXiv preprint arXiv:1812.07035}, 2020.

\end{thebibliography}

\end{document}